\documentclass{ecai} 

\usepackage{latexsym}
\usepackage{amssymb}
\usepackage{amsmath}
\usepackage{amsthm}
\usepackage{booktabs}
\usepackage{enumitem}
\usepackage{graphicx}
\usepackage{color}
\usepackage{bm}

\newcommand{\BibTeX}{B\kern-.05em{\sc i\kern-.025em b}\kern-.08em\TeX}

\newcommand{\bp}{{\bf p}}

\begin{document}


\begin{frontmatter}


\paperid{1153} 


\title{Theoretical Proportion Label Perturbation \\
for Learning from Label Proportions in Large Bags }


\author[A]{\fnms{Shunsuke}~\snm{Kubo}}
\author[A]{\fnms{Shinnosuke}~\snm{Matsuo}}
\author[A,D]{\fnms{Daiki}~\snm{Suehiro}}
\author[B]{\fnms{Kazuhiro}~\snm{Terada}}
\author[B]{\fnms{Hiroaki}~\snm{Ito}}
\author[C]{\fnms{Akihiko}~\snm{Yoshizawa}}
\author[A]{\fnms{Ryoma}~\snm{Bise}}
\address[A]{Kyushu University}
\address[B]{Kyoto University Hospital}
\address[C]{Nara Medical University}
\address[D]{RIKEN AIP}



\begin{abstract}
Learning from label proportions (LLP) is a kind of weakly supervised learning that trains an instance-level classifier from label proportions of bags, which consist of sets of instances without using instance labels.
A challenge in LLP arises when the number of instances in a bag (bag size) is numerous, making the traditional LLP methods difficult due to GPU memory limitations.
This study aims to develop an LLP method capable of learning from bags with large sizes. 
In our method, smaller bags (mini-bags) are generated by sampling instances from large-sized bags (original bags), and these mini-bags are used in place of the original bags. However, the proportion of a mini-bag is unknown and differs from that of the original bag, leading to overfitting.
To address this issue, we propose a perturbation method for the proportion labels of sampled mini-bags to mitigate overfitting to noisy label proportions. This perturbation is added based on the multivariate hypergeometric distribution, which is statistically modeled. Additionally, loss weighting is implemented to reduce the negative impact of proportions sampled from the tail of the distribution.
Experimental results demonstrate that the proportion label perturbation and loss weighting achieve classification accuracy comparable to that obtained without sampling.
Our codes are available at https://github.com/stainlessnight/LLP-LargeBags.
\end{abstract}

\end{frontmatter}


\section{Introduction}
\label{sec:intro}
\begin{figure}[t]
\begin{center}
    \includegraphics[width=0.83\linewidth]{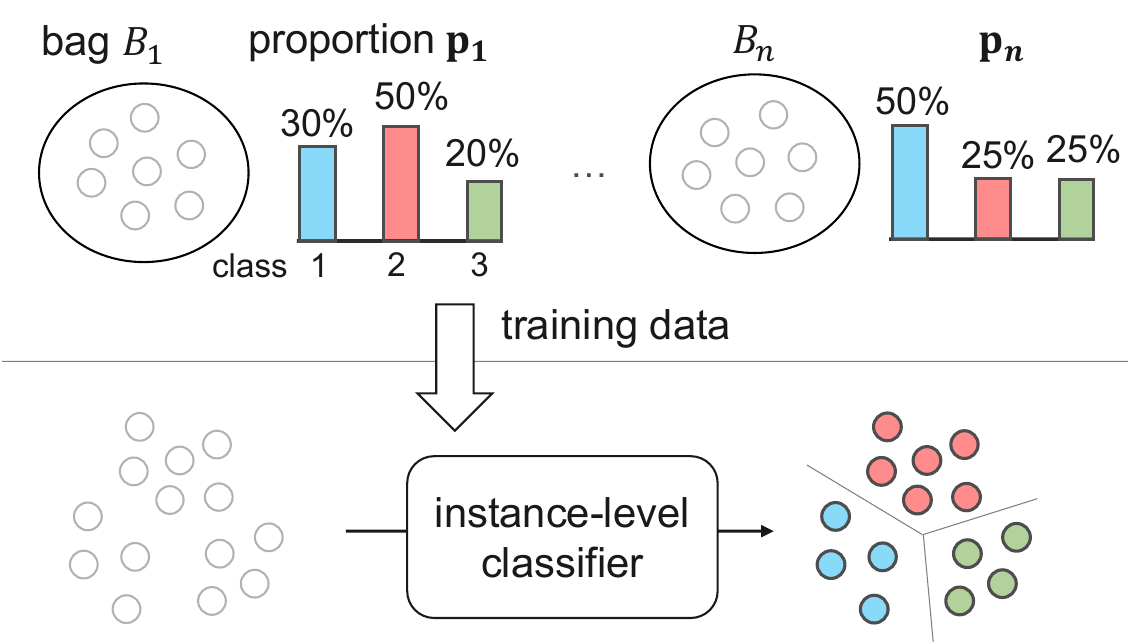}
\end{center}
\vspace{5pt}
\caption{Learning from label proportions (LLP). An instance-level classifier is trained using bag-level labels (label proportion in a bag), where a bag consists of a set of instances and the labels of individual instances are unknown.}
\vspace{20pt}
\label{fig:llp}
\end{figure}

Learning from label proportions (LLP) is one of the promising problems of weakly supervised learning~\cite{ardehaly2017co,med_image,remote_sensing,dulacarnold2019deep,liu2019llpgan,election_pred, tokunaga2020negative,pmlr-v129-tsai20a,yang2021two}. 
Figure \ref{fig:llp} illustrates the problem setup of LLP.
In LLP, label proportions of sets (referred to as bags) of instances (e.g., images) are given as training data, but individual instance-level labels are unknown. The objective of LLP is to train an instance-level classifier leveraging the label proportions provided at the bag level.
Solving LLP is more challenging compared to supervised learning setups because bag-level labels (class proportion) are only given with unknown instance-level labels, making it difficult to compute a loss for each instance.
\par
Hence, LLP is suitable for situations where direct class labels are unavailable but aggregate information is provided.
While LLP is closely related to multiple-instance learning (MIL), which utilizes bag-level class labels as training data, their objectives differ: LLP aims to train an instance-level classifier, whereas MIL aims to train a bag-level classifier.
\par
LLP is beneficial in a variety of applications. It is particularly useful in scenarios where the protection of individual instance labels' privacy is imperative, in situations where annotating instances is challenging, or when there is a desire to reduce the costs associated with annotation. Examples of its application include election prediction \cite{election_pred}, medical image analysis \cite{med_image,matsuo2024lplp,tokunaga2020negative}, and remote sensing \cite{remote_sensing}. 
These fields benefit from LLP's ability to work with aggregate label information.

\par
Many methods for LLP have been proposed. The recent trend in LLP is to train a Deep Neural Network (DNN) using a proportion loss, originally proposed by Ardehaly et al.~\cite{ardehaly2017co}. The proportion loss calculates a bag-level cross-entropy between the ground truth and the predicted proportion, which is the average of the probability outputs of individual instances of each bag. Several methods extend the proportion loss by introducing regularization terms or pre-training techniques~\cite{dulacarnold2019deep,liu2019llpgan,pmlr-v129-tsai20a,yang2021two}.
\par
These methods face limitations, particularly when dealing with large bags (i.e., bags containing a large number of instances).
To train a model using the loss calculated from the estimated probabilities of all instances within a bag, these instances should be simultaneously inputted into a batch, maintaining all gradients of the loss during a training iteration.
However, loading all instances from a large bag into a batch is challenging due to GPU memory limitations.
\par
This situation of large-size bags often occurs in real-world applications where statistical information (proportions) is provided for extensive datasets\footnote{Note that the bag size in LLP is given as training data, so users cannot control it.}.
For example, consider window-wise long-term video classification, where each video is represented as a large bag containing numerous instances corresponding to individual windows.
Another example is whole slide image (WSI) segmentation, where cancer grading relies on the proportions of tumor subtypes, such as chemotherapy~\cite{travis2020iaslc}, the PDL1 positive rate~\cite{Liu2021,okuoT2023,roach2016development,Widmaier2020}, and lung tissue diagnosis after surgery~\cite{Dacic2021,moreira2020grading}, resulting in recorded label proportions during diagnosis.
WSIs are typically large images, often spanning ten thousand pixels in width and height. To manage them effectively, images are divided into smaller patch images. Consequently, a WSI can be considered as a bag, with the set of patch images within a WSI representing the instances.
In such cases, all instances in a bag cannot be inputted in a batch.
\par
A naive sampling method can address GPU memory limitations when dealing with large-size bags in LLP. This method involves randomly sampling instances (creating a mini-bag) from the original large bag, enabling simultaneous input of all instances in the mini-bag into a batch.
In this approach, the original large bag's label proportions are used as the mini-bag proportion's supervision.
However, due to random sampling, the original bag's proportion used as supervision for the mini-bag may differ from the ground truth proportion of the mini-bag. For example, consider a case where 3 instances are randomly sampled from a set of 10 instances (a bag) containing 7 red balls and 3 blue balls, resulting in a proportion of 0.7 for red balls. If three red balls are randomly sampled, the proportion becomes 1.0, which differs from the original proportion of 0.7.
This proportion shift can be considered as label noise, and it leads to overfitting to incorrect label proportions. 
\par
In this paper, we propose a Learning from Label Proportion (LLP) method that is effective for scenarios with large numbers of instances within a bag. Initially, we investigate the relationship between proportion error and sample size to mitigate label noise when forming mini-bags. We also empirically show its impact on instance-level classification accuracy.
Subsequently, we statistically analyze the distribution of label proportions in mini-bags sampled from the original bags. This distribution is theoretically modeled using the multivariate hypergeometric distribution, forming the basis of our method.

Our approach involves perturbing the supervision of mini-bag proportions by randomly sampling the proportion labels according to the multivariate hypergeometric distribution. This perturbation varies across training iterations, with different proportions randomly selected in each iteration, thereby preventing overfitting to noisy label proportions. 
Sometimes, there is a possibility of randomly sampling proportions at the tail of the distribution, which can significantly deviate from the ground truth of the mini-bag and potentially impact the learning negatively.
We introduce loss weighting according to this distribution to mitigate the adverse effects of large noisy labels.
The effectiveness of our proposed method is validated through experiments on CIFAR-10, SVHN, and a pathological image dataset. Our method outperformed comparative methods across various scenarios, demonstrating its robustness and efficacy in handling changing sample sizes.

The contributions of this study are summarized as follows:
\begin{itemize}
    \item We investigate the relationship between the proportion error and the sample size, and we empirically show its bad impact on the instance-level classification accuracy.
    \item We have theoretically modeled the distribution of label proportions in a mini-bag sampled from an original bag as a multivariate hypergeometric distribution.
    \item We propose proportion label perturbation for LLP by sampling proportion labels along the multivariate hypergeometric distribution. This perturbation avoids overfitting.
    \item We further propose a loss weighting mechanism based on the probability mass function of the hypergeometric distribution for label proportions. 
    \item The effectiveness of these proportion label perturbation and loss weighting based on the distributions is demonstrated in various situations.
\end{itemize}

\section{Related Work}
\label{sec:relatedwork}
\noindent
{\bf Learning from Label Proportions:}
A commonly employed method in LLP is a proportion loss (PL), which is calculated as the bag-level cross-entropy between the ground truth label proportion and the predicted proportion (refer to Section 3 for details). Additionally, various approaches have been introduced to enhance the learning process.
For example, LLP-VAT~\cite{pmlr-v129-tsai20a} incorporates Virtual Adversarial Training (VAT) along with the proportion loss. VAT aims to train the model so that the posterior probabilities remain consistent between an original image and the slightly perturbed image.
LLP-GAN~\cite{liu2019llpgan} introduces Generative Adversarial Networks (GANs) into the framework. In this approach, class classification at the instance level is performed on a discriminator trained via GAN, and the predicted posterior probabilities are utilized in the proportion loss.
The Two-Stage for LLP~\cite{liu2021two} involves first training a class classifier using the proportion loss and then providing pseudo-labels at the instance level, constrained by the label proportions. Subsequently, the classifier trained with the proportion loss is further trained using these pseudo-labels. MixBag~\cite{asanomi2023mixbag} is a bag-level data augmentation technique for LLP, which introduces a confidence interval loss based on the distribution of proportions in sampled mix-bags.
A common challenge faced by these methods is their limited applicability to large-sized bags, as the proportion loss necessitates inputting all instances in a batch.

Some methods take a pseudo-labeling approach, which assigns pseudo-labels to each instance~\cite{matsuo2023opl,Zhang2022llpfc}. LLPFC~\cite{Zhang2022llpfc} utilizes the noise transition matrix, estimated based on proportion, for pseudo-labeling. Online Pseudo-Label Decision (OPL)~\cite{matsuo2023opl} introduces a regret minimization framework, which is one of the theories for online decision-making. These methods do not require inputting all instances into a network jointly in a batch. Therefore, they have an advantage for large bag sizes. However, these methods assume that the distributions of each class are roughly separated in the initial feature spaces, as the proportion information cannot be utilized for initial representation learning. This assumption does not often hold in real-world applications; these methods may not perform well. The effectiveness of these methods is demonstrated only using toy datasets in the papers; hence, their performance on real-world datasets needs further investigation.

We propose a solution by handling bags of reduced size, where the representation learning can be done using proportion information, thereby addressing this limitation.

\vspace{0.5\baselineskip}
\noindent
{\bf Learning with Noisy Labels:}
The necessity for robust learning methods in Deep Neural Networks (DNNs) under the presence of incorrect supervisory labels has become increasingly apparent. DNNs possess a remarkable capacity for memorization~\cite{memorization}, demonstrated by their ability to reduce training error to zero even when trained with randomly assigned labels. While showcasing the flexibility of DNNs, this capability simultaneously underscores their vulnerability to overfitting on incorrectly labeled data, impeding their ability to generalize effectively from such datasets. The challenge of accumulating high-quality data samples and their corresponding class labels is a significant issue in real-world applications, often resulting in datasets contaminated with erroneously labeled instances ~\cite{clothes1m}.
\par
In response to these challenges, several robust learning methodologies have been proposed, capable of maintaining resilience even in the presence of label noise. Song et al~\cite{nl_survey} have categorized these methods into five distinct approaches: Robust Architecture~\cite{Chen2015, clothes1m}, Robust Regularization~\cite{pmlr-v97-hendrycks19a, pereyra2017regularizing}, Robust Loss Function~\cite{ma2020}, Loss Adjustment~\cite{pat2017, wang2017im}, and Sample Selection~\cite{han2018,wei2020}.
Robust Architecture involves modifications to the network structure, such as inferring noise transition matrices and incorporating noise adaptation layers within the network.
Robust Regularization employs techniques like dropout, weight decay, data augmentation, and mini-batch SGD to regularize DNN weights.
Robust Loss Function employs loss functions specifically designed to mitigate overfitting caused by incorrect labels. Cross-entropy loss is prone to overfitting caused by incorrect labels, but alternatives such as Mean Absolute Error (MAE) and Reverse Cross-Entropy are considered more robust.
Loss Adjustment~\cite{pat2017, wang2017im} refers to the modification of loss values through methods like noise transition matrix-based loss correction, loss reweighting, and incorrect label correction.
Sample Selection~\cite{han2018,wei2020} differentiates between incorrect and correct labels, training separate networks with correctly labeled instances or employing semi-supervised learning techniques to discard incorrectly labeled data. Methods for identifying incorrect labels include the low-loss trick and examining previous prediction outcomes of data samples.
\par
Among the various strategies developed to enhance robustness in learning with noisy labels, Loss Reweighting ~\cite{wang2017im,zhang2021dual} has emerged as a significant approach. This technique involves adjusting individual samples' loss contributions based on their labels' correctness, effectively reducing the loss attributed to samples with incorrect labels while amplifying the loss for those with correct labels. By recalibrating the loss in this manner, the influence of incorrectly labeled data samples on the learning process is minimized, thereby mitigating the adverse effects of label noise on model performance.
\par
In relation to these researches, our study focuses on the discrepancies between the label proportions assigned to mini-bags generated through instance sampling and the true label proportions of these mini-bags. We interpret learning with inaccurately assigned label proportions as Learning with Noisy Labels. We introduce a loss-reweighting approach by considering the occurrence probabilities of label proportions sampled from a hypergeometric distribution as a measure of label proportion reliability. This methodology aims to mitigate the adverse effects of label noise, thereby enhancing the robustness and generalization capabilities of DNNs trained under such conditions.

\section{Preliminary: LLP and proportion loss}
In LLP, the training data consists of $n$ labeled bags, denoted as $\mathcal{B} = \{B^i, \bp^i\}_{i=1}^n$, where each bag $B^i$ comprises a set of instances represented as $B^i=\{x_j^i\}_{j=1}^{N^i}$, where $N^i$ is the bag size. The vector $\bp^i=(p^i_1,\ldots,p^i_c,\ldots,p^i_C)^\mathsf{T}$ contains the label proportions, indicating the ratios of the number of instances belonging to each class in a bag (i.e., $\sum_{c=1}^C p^i_c =1$ and $0 \leq p_c \leq 1$). The class label for each instance is unknown.
The objective of LLP is to train an instance-level classifier that estimates the class label of each instance using only the bag-level proportion labels without access to instance-level labels.

In the realm of LLP, numerous studies \cite{ardehaly2017co, asanomi2023mixbag, liu2019llpgan, liu2021two, pmlr-v129-tsai20a} have integrated the concept of the proportion loss \cite{ardehaly2017co}.
When provided with training data containing label proportions $\mathcal{B}$, the standard proportion loss is calculated as the bag-level cross-entropy between the ground truth label proportion $\bp^i$ and the predicted proportion $\hat{\bp}^i$. The predicted proportion is computed as the average of the output class probabilities of the instance classifier $f$ for all instances within a bag. The proportion loss is defined as:
\begin{align}
    \ell_{\mathrm{prop}}(B^i, \bp^i, f) = - \sum_{c=1}^{C} p_c^i \log \hat{p}_c^i, \\
    \hat{p}_c^i = \frac{1}{N^i} \sum_{j=1}^{N^i} f(x_j^i)_c,
\end{align}
where $f(x_j^i)_c$ represents the output probability (called confidence) that instance $x_j^i$ belongs to class $c$.

A notable challenge in applying the proportion loss within the LLP framework is the necessity to load all instances contained within a bag into the GPU, which presents substantial difficulties as the size of the bag increases. Recognizing this limitation, this paper aims to explore and develop methods that enable the application of the proportion loss in scenarios involving large bags. By addressing this challenge, the proposed approaches seek to significantly enhance the scalability and applicability of LLP models, thereby broadening the potential for their implementation in various domains.

\section{Preliminary experiment: Impact by sample size}
Sampling from the original bag is a simple method 
to tackle the memory issue in LLP. 
Naively, 
a set of sampled instances (mini-bag) from an original bag is used as a bag, and the original bag's proportions are used as the bag-level supervision in training.
However, the mini-bag's proportion may differ from that of the original bag, and the differences (gap) of the distribution 
depends on the original bag size, 
sample size, and proportion. 
We consider 
that the gap increases with a smaller sample size.
To 
assess the impact of sample size from a large bag for classification accuracy in LLP, we conducted experiments using the naive method with changing sample size.

\begin{figure}[t]
    
 \begin{center}
    \begin{minipage}[b]{0.48\linewidth}
        \centering
        \includegraphics[width=\linewidth]{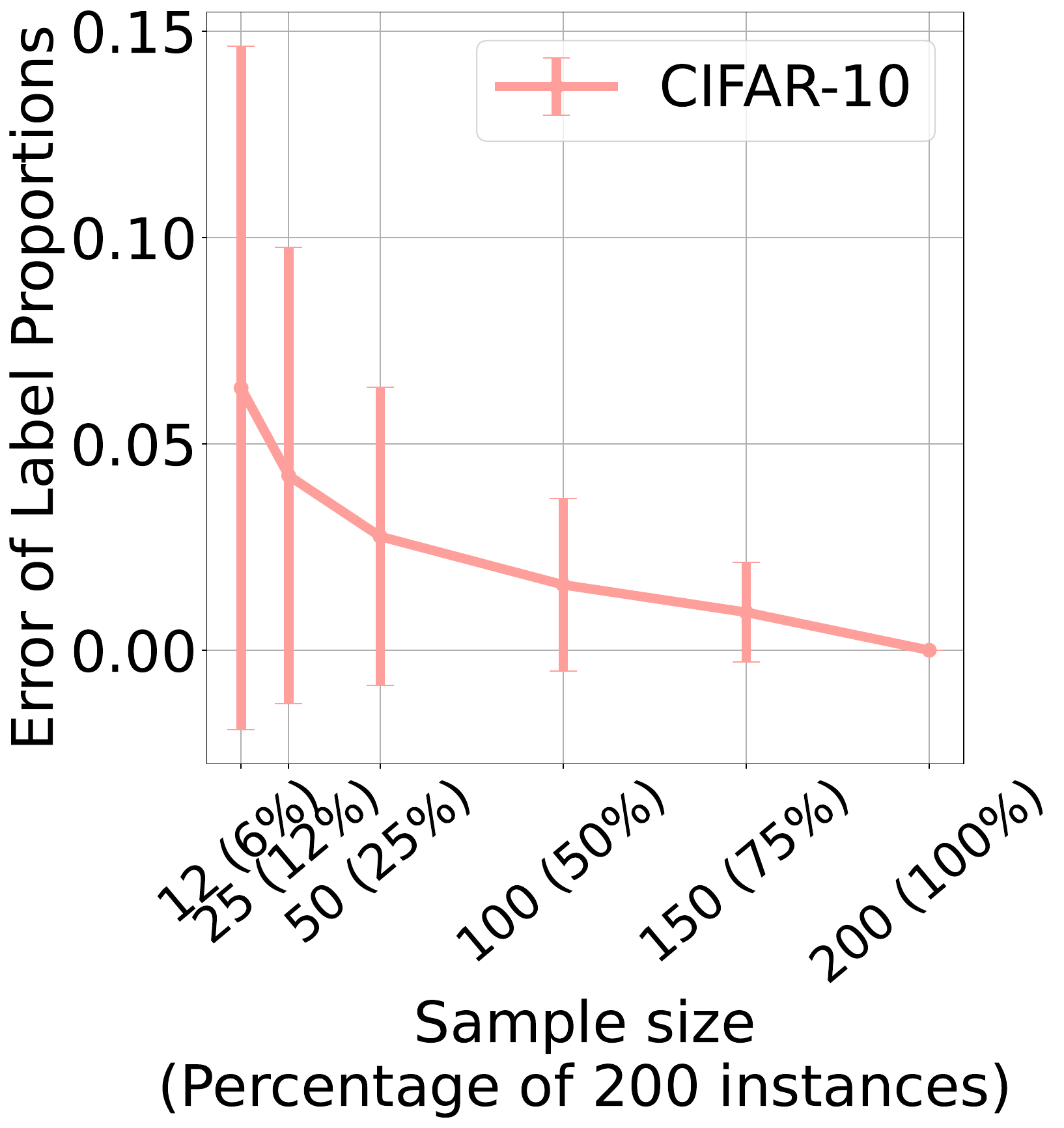}
    \end{minipage}
    \hfill
    \begin{minipage}[b]{0.465\linewidth}
        \centering
        \includegraphics[width=\linewidth]{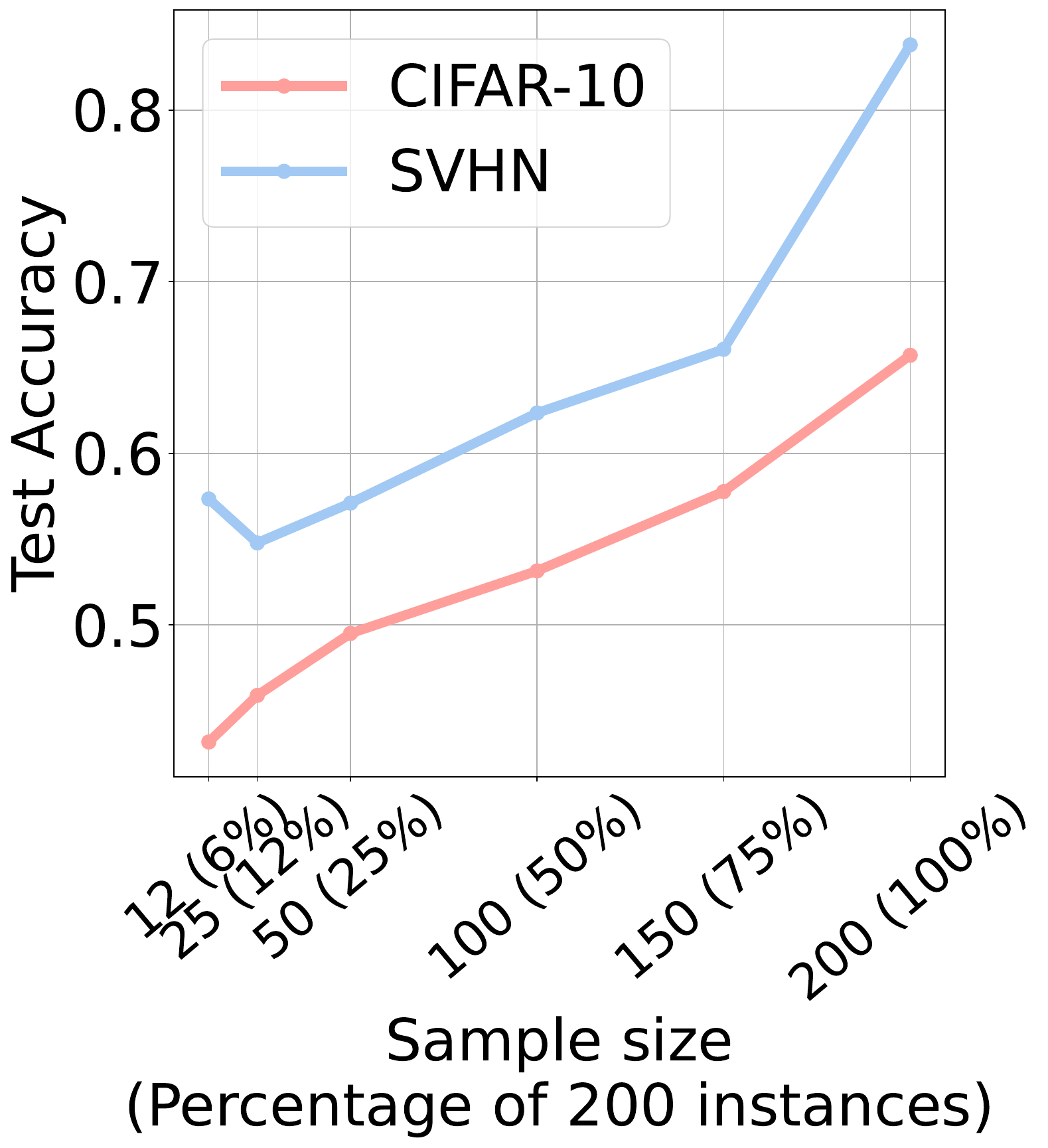}
    \end{minipage}
 \end{center}
 \vspace{-3mm}
 \vspace{5pt}
\caption{(Left): Mean absolute error (MAE) of label proportions between the mini-bag's ground truth and the original bag when changing the sample size. The error bar shows the standard deviation. (Right): Classification accuracy of each sample size. Red and blue lines indicate the results on CIFAR-10 and SVHN, respectively. Since the MAEs are mostly the same on both datasets, SVHN is omitted in the left figure.}
\label{fig:pre_results}
\vspace{20pt}
\end{figure}

\vspace{0.5\baselineskip}
\noindent
{\bf Experimental setup:}
We used a ResNet-18~\cite{resnet18} as the network backbone and used a proportion loss for training it. We employed the original bag's label proportions as the supervised proportion for the sampled mini-bags to compute proportion loss for mini-bags.

We used the CIFAR-10 dataset~\cite{cifar10}, which is widely used as a standard dataset in LLP papers~\cite{asanomi2023mixbag,liu2019llpgan, matsuo2023opl, pmlr-v129-tsai20a, Zhang2022llpfc}. To create a bag, we followed previous studies~\cite{asanomi2023mixbag, matsuo2023opl, Zhang2022llpfc}; the proportions of each class were chosen randomly from a Gaussian distribution. The selected proportion was employed as the bag's label. Subsequently, samples were randomly chosen from each class, with the number of samples for each class corresponding to the class's proportion. We constructed 250 bags, where each bag contains 200 instances, and the number of classes is 10.
For evaluation, we employed the test data provided with the CIFAR-10 dataset to assess instance-level classification accuracy.
Metrics were calculated using a five-fold cross-validation approach.
We evaluated the instance-level accuracy for test data with changing sample sizes: 12, 25, 50, 100, 150, and 200.

\vspace{0.5\baselineskip}
\noindent
{\bf Proportion errors in sampled mini-bags:}
Figure~\ref{fig:pre_results}(Left) shows the mean absolute error (MAE) and the standard deviation (SD) of the proportion gaps between the original bag and mini-bag when changing the sample size of mini-bags, where the horizontal axis is the sample size, and the vertical axis shows the MAE.
The sample size of 200 indicates that all instances in a bag are sampled; the mini-bag is the same as the original bag. Therefore, the error is 0.
As the sample size decreases, the average MAE increases. It is common in the statistics view; as the size of a sample or dataset increases, the average of the data will tend to converge to the population's true mean.

\vspace{0.5\baselineskip}
\noindent
{\bf Instance-level accuracy when changing sample size:}
The proportion gap between the original bag and the mini-bag may negatively affect the training in LLP. We next evaluated the instance-level accuracy when changing the sample size of mini-bags.

Figure~\ref{fig:pre_results}(Right) shows the results on two datasets: CIFAR-10 and SVHN.
As the sample size decreases, the accuracy decreases. The proportion gaps caused by sampling can be considered as noise of label proportions, and thus, training forces the model to overfit to noisy labels. Therefore, the smaller sample size caused the larger error, and it led to decreasing accuracy.

\begin{figure}[t]
 \begin{center}
    \includegraphics[width=0.9\linewidth]{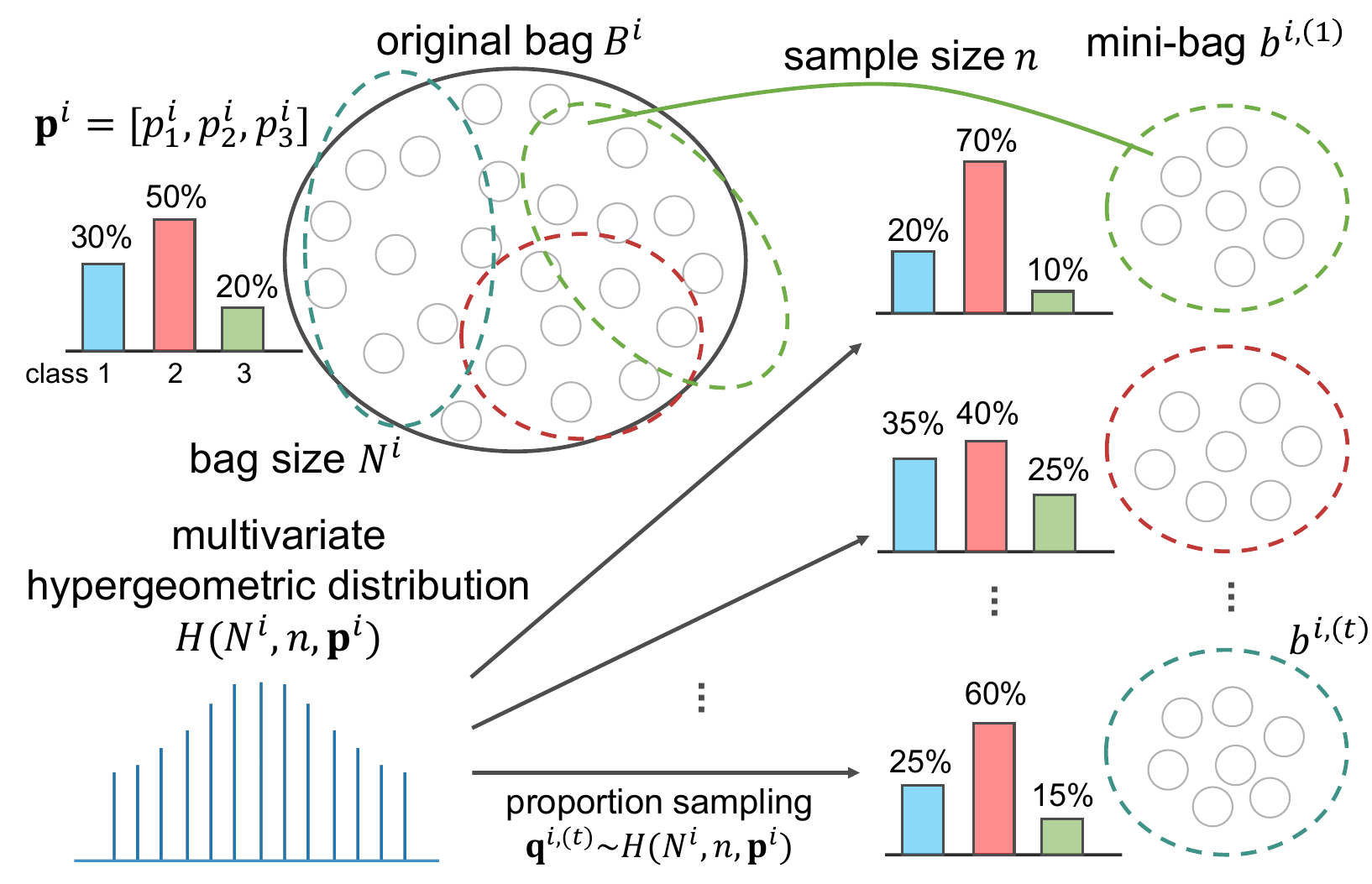}
    \vspace{5pt}
    \caption{Overview of the perturbation for the proportion of mini-bags. Random sampling generates a mini-bag from the original bag in each iteration. The supervision for the proportion of the mini-bags is also randomly determined along with the multivariate hypergeometric distribution. This perturbation can mitigate overfitting to the noisy proportion.}
    \label{fig:overview}
 \end{center}
 \vspace{10pt}
\end{figure}

\section{Theoretical proportion label perturbation for LLP when sampling from large bags}

When the supervised proportion label contains noise, the main reason for reducing the accuracy can be considered as the noisy label was fixed on each iteration. The model is trained to produce the noisy label in each iteration, leading to overfitting into the noise. 

We propose a perturbation method for the proportion labels of sampled mini-bags from a large original bag in LLP to mitigate overfitting to a noisy label of proportions. Instead of training a network with fixed noisy labels, we apply different perturbations to the proportion labels for each mini-bag in different iterations as shown in Figure~\ref{fig:overview}.
Due to the perturbation, the model does not converge to the noisy label, while it gets closer to the optimal solution for instance-level classification.

We theoretically model the distribution of label proportions in a mini-bag sampled from an original bag. The distribution can be modeled by multivariate hypergeometric distribution.
Our method utilizes this multivariate hypergeometric distribution to apply perturbations to the proportion labels and to weight the loss accordingly.

\begin{figure}[t]
 \begin{center}
    \includegraphics[width=0.95\linewidth]{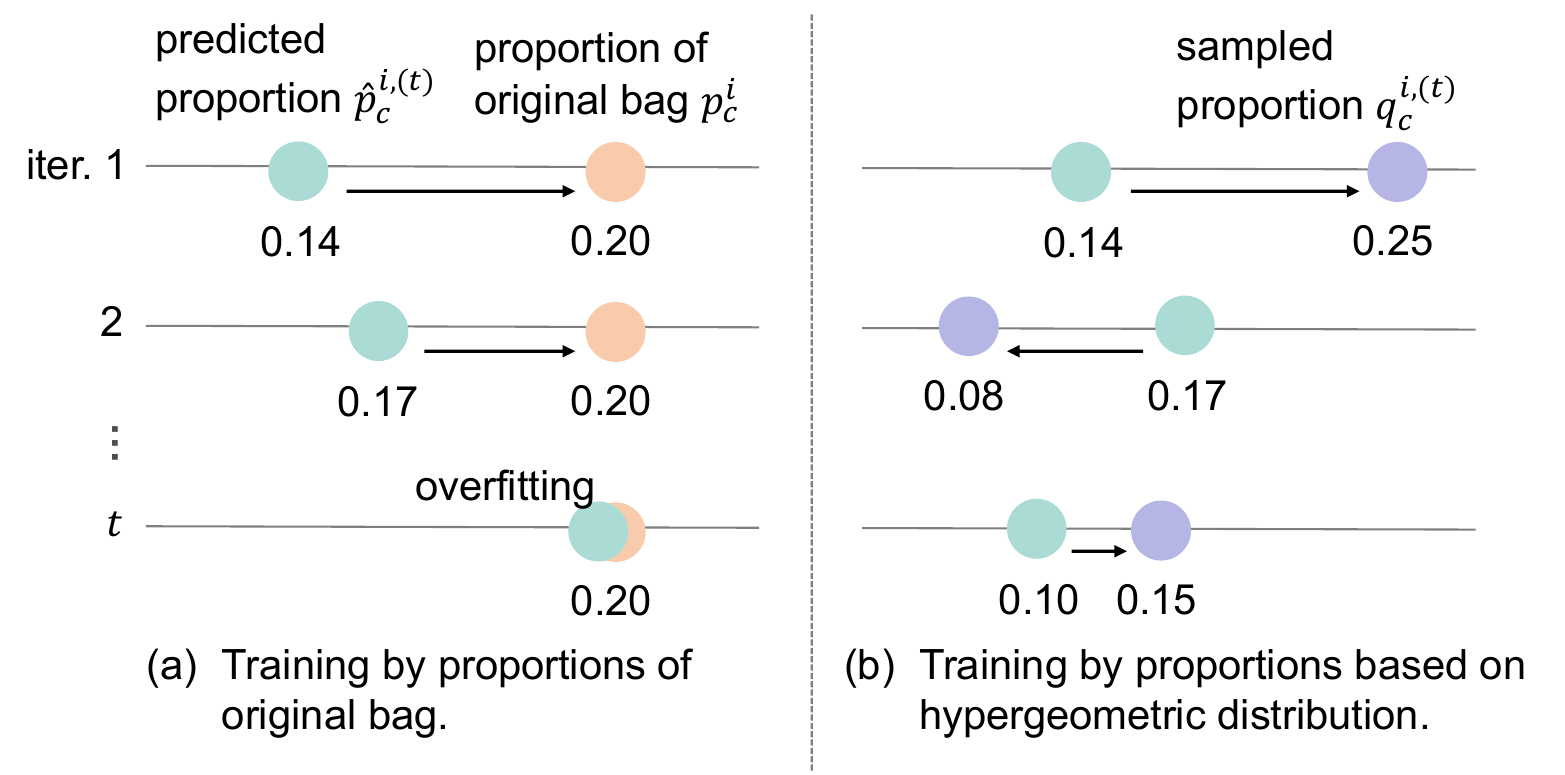}
    \vspace{5pt}
    \caption{Direction of gradients at each iteration. 
    (a) A case when using original bag label proportions. The estimated proportion converged to the noisy proportion (overfitting). (b) A case when using label proportions sampled from the multivariate hypergeometric distribution. It does not converge to the noisy proportion due to perturbation.
    }
    \label{fig:gradient_direction}
 \end{center}
 \vspace{10pt}
\end{figure}

\subsection{Modeling the distribution of label proportions of mini-bags through sampling}

In this section, we first statistically investigate the distribution of label proportions in a mini-bag sampled from an original bag.
To simplify the discussion, let us first consider a case of binary classification, where an instance belongs to either the positive or negative class.
Given a set of $N$ instances in an original bag that contains $K$ positive instances, $n$ instances are sampled from the original bag. In this case, it is known that the probability mass function of the number of positive instances follows the hypergeometric distribution.

This can be extended to a multi-class case. Given a bag whose label proportion is $\mathbf{p} = [p_1, p_2, \ldots, p_C]$ and size is $N$ (i.e., the number of instances belonging to class $c$ is $K_c=N p_c$), sampling $n$ instances from the original bag.
The random variable $\mathbf{X}=[X_1, X_2, \ldots, X_C]$, which indicates the number of instances in each class, follows a multivariate hypergeometric distribution $H(N,n,\mathbf{p})$ and it's mass function $P_H$ is described as:
\begin{equation}
P_H(X_1 = k_1, \ldots, X_C = k_C; N, n, \mathbf{p}) = \frac{\prod_{c=1}^{C}\binom{K_c}{k_c}}{\binom{N}{n}},
\end{equation}
where $\binom{a}{b}$ indicates a binomial coefficient.

Based on the multivariate hypergeometric distribution $H(N,n,\mathbf{p})$, we can estimate the probability of occurrence of a label proportion $\mathbf{\Tilde{p}} = [\Tilde{p_1}, \Tilde{p_2}, \ldots, \Tilde{p_C}]$ of the sampled mini-bag, where the number of $c$ class instances $k_c = n \Tilde{p_c}$.
The proposed method uses this multivariate hypergeometric distribution for perturbation and loss weighting.

\begin{table*}[t]
\centering
\fontsize{8pt}{10pt}\selectfont
\begin{tabular}{l|cccccc|cccccc}
\hline
Dataset & \multicolumn{6}{c|}{CIFAR-10} & \multicolumn{6}{c}{SVHN} \\ \hline
 Sample size & 12 & 25 & 50 & 100 & 150 & 200 & 12 & 25 & 50 & 100 & 150 & 200 \\
\hline
PL & 0.4318 & 0.4589 & 0.4951 & 0.5315 & 0.5777 & 0.6571 & 0.5734 & 0.5478 & 0.5710 & 0.6235 & 0.6606 & 0.8380 \\ \hline
Ours & \bf{0.5448} & \bf{0.5978} & \bf{0.6401} & \bf{0.6589} & \bf{0.6573} &      & \bf{0.7820} & \bf{0.8063} & \bf{0.8464} & \bf{0.8669} & \bf{0.8719} & \cr
Ours w/o LW & 0.5145 & 0.4991 & 0.5460 & 0.5960 & 0.6244 &      & 0.7518 & 0.6626 & 0.6843 & 0.7393 & 0.7464 & \cr
\hline
LLP-VAT & 0.4577 & 0.4887 & 0.5279 & 0.5448 & 0.5415 & 0.5394 & 0.6393 & 0.5891 & 0.6601 & 0.5788 & 0.5389 & 0.5235 \\ \hline
Ours+LLP-VAT & \bf{0.5692} & \bf{0.6089} & \bf{0.6350} & \bf{0.6367} & \bf{0.5887} &       & \bf{0.7990} & \bf{0.8279} & \bf{0.8505} & \bf{0.8490} & \bf{0.7739} & \cr
Ours+LLP-VAT w/o LW & 0.5397 & 0.5343 & 0.5515 & 0.5631 & 0.5538 &       & 0.7684 & 0.7249 & 0.6871 & 0.6496 & 0.5579 & \cr
\hline
\end{tabular}
\vspace{5pt}
\caption{Accuracy of each method when changing the sample size.}
\label{tab:toy}
\vspace{20pt}
\end{table*}

\subsection{Proportion label perturbation using multivariate hypergeometric distribution}
In our method, we use a standard proportion loss to train a network, which sums the output estimation by a convolutional neural network (CNN) for individual instances (images) in a bag to estimate the label proportion and calculates the loss between the ground truth and estimation of label proportions.

As previously mentioned, due to GPU memory constraints, a mini-bag is created from an original bag through random sampling. Consequently, the ground truth of the mini-bag's proportion remains unknown.
If we naively use the original bag's label proportion as the mini-bags supervision, the supervision contains noise (i.e., shift from its ground truth); in particular, the case when the sample size $n$ is much smaller than the original size $N$. 
If we use the noisy proportion label with fixed in the entire training, the network tends to overfit the noisy supervision as shown in Figure~\ref{fig:gradient_direction} (a), which converged to the noisy proportion.

To mitigate overfitting, we add perturbation for the supervision of the mini-bag's proportion sampled from bag $B^i$ in each iteration $t$ during training, where at different iterations, a different mini-bag $b^{i,(t)}$ is also randomly sampled from bag $B^i$.
The perturbation is implemented by randomly sampling the supervised proportion from the multivariate hypergeometric distribution, defined as follows:
\begin{equation}
\mathbf{q}^{i,(t)} \sim H(N^i,n,\mathbf{p}^i),
\end{equation}
where $\mathbf{q}^{i,(t)}$ is the supervision of the proportions of the mini-bag on the $t$-th iteration in training. 
Note that as discussed in the above section, the ground truth of mini-bags proportions $\mathbf{\Tilde{p}}^{i,(t)}$, which is unknown, also follow the multivariate hypergeometric distribution $H(N,n,\mathbf{p}^i)$.

Figure~\ref{fig:gradient_direction}(b) illustrates the gradient direction in each iteration during the training of our method.
Our method provides different perturbations in each iteration, which helps to mitigate overfitting.
Although the proportion loss does not converge due to perturbation, the representation learning for the feature extractor progresses well because the perturbed proportion follows the multivariate hypergeometric distribution modeled statistically, which depends on the sample size and the original bag's proportion.
This enables improving the instance-level accuracy.

\subsection{Weighted loss based on multivariate hypergeometric distribution}

The randomly sampled proportion may sometimes contain significant noise, which can deviate significantly from the ground truth of the mini-bags proportion.
Although such large noise is rarely sampled due to the very small occurrence probability in $H(N^i,n,\mathbf{p}^i)$, it may harm training.

To avoid this, we introduce loss weighting based on the mass function of the multivariate hypergeometric distribution ($w_i = P_H(N^i,n,\mathbf{p}^i)$): a weight takes a high value when the randomly sampled population is around the original bag's one and a small value when it is around the distribution tail.
In the implementation, the weights are normalized by dividing by the median of the batch.
The weighted loss function is formalized as follows:
\begin{eqnarray}
\ell_{\mathrm{prop}}^w = \sum_{i \in \textbf{Batch}^{(t)}}{w_i \ell_{\mathrm{prop}}(\mathbf{\hat{p}}^{i,(t)},\mathbf{q}^{i,(t)})},\\
\hat{p}_c^{i,(t)} = \frac{1}{n} \sum_{x \in b^{i,(t)}}{f(x)_c},
\end{eqnarray}
where $\mathbf{\hat{p}}^{i,(t)}$ is the estimated proportion of the mini-bag by the trained network model $f$ in iteration $t$.
$\mathbf{p}^i$ is the proportion of the corresponding original bag $B^i$.
$\textbf{Batch}^{(t)}$ is the set of indices of the bags in a batch at iteration $t$.
The estimated proportion $\hat{p}_c^{i,(t)}$ of the input bag $B^i$ for class $c$ is the mean of the estimated confidence $f(x)_c$ for class $c$ of instances $x \in b^{i,(t)}$.
The weight indicates the normalized probability on $H$ in a batch.
Using this weighted loss, the instance classifier model $f$ is trained.
The trained network can estimate the class of individual instances without inputting a bag.

\section{Experiments on public data}
\label{sec:experiments}

We first evaluated the performance of our method using public datasets, following the evaluation approach of many related works on LLP.

\vspace{0.5\baselineskip}
\noindent
{\bf Dataset:}
We evaluated the proposed method using CIFAR-10~\cite{cifar10} and SVHN~\cite{svhn}. CIFAR-10 and SVHN consist of 10 classes. The original bags for CIFAR-10 and SVHN were created under the same conditions as in the preliminary experiments.

\vspace{0.5\baselineskip}
\noindent
{\bf Experimental setup:}
As the network, we used ResNet-18~\cite{resnet18}, which was pre-trained on ImageNet~\cite{imagenet}. The batch size was set to 8 bags, and the learning rate was 0.0001.
The models were trained for a maximum of 50 epochs, and the model with the lowest proportion loss on the validation data was selected. The train data and validation data were split in a 4:1 ratio. The training was conducted 5 times, and the evaluation metric was the average accuracy of the test data across these 5 runs.

\vspace{0.5\baselineskip}
\noindent
{\bf Ablation study:}
We conducted an ablation study to verify the effectiveness of the proposed method's perturbation approach for proportion labels and loss weighting based on the multivariate hypergeometric distribution.
Specifically, we evaluated the proposed method compared with two methods as an ablation study: 1) a baseline method using proportion loss (PL)~\cite{ardehaly2017co}, which directly used the original bag's proportion for the supervision of the sampled mini-bag in the same manner as in the preliminary experiments. Note that different mini-bags are generated on different iterations for fair comparison.
2) the proposed method (Ours) that gives perturbation for proportion labels and uses loss weighting. 3) the proposed method without using loss weighting (Ours w/o LW). 

Table~\ref{tab:toy} shows the instance-level accuracy for each method when changing the sample size. As shown in the preliminary experiment (Section 4), as the sample size decreases, the accuracy decreases due to the noise by sampling on both datasets. Ours w/o LW improved the accuracy for all cases from the baseline by adding perturbation for proportion labels in each iteration based on the multivariate hypergeometric distribution.
The proposed method (Ours) further improved the accuracy by weighting the loss based on the distribution. These results indicate the effectiveness of both the perturbation approach and loss weighting.




\vspace{0.5\baselineskip}
\noindent
{\bf Effectiveness on different backbone method:}
To show the scalability of our method, we also applied our method to a different baseline method for LLP. Specifically, we introduce the perturbation approach and loss weighting into LLP-VAT~\cite{pmlr-v129-tsai20a}, which introduces the consistency regularization into the proportion loss. The results are shown in Table~\ref{tab:toy}. This method also decreased the accuracy when the sample size decreased. Our method consistently improved accuracy across all sample sizes. Our method is applicable to any LLP method that computes a bag-level loss, such as the proportion loss. The experimental results demonstrate the potential for improved accuracy with our proposed method across various LLP methods when the bag size is large.

\vspace{0.5\baselineskip}
\noindent
{\bf Effectiveness of the proposed method for mitigating overfitting:}
To demonstrate the effectiveness of our method in mitigating overfitting, we analyzed the loss curve and the instance-level accuracy of the training data.
In LLP, the primary objective of the loss function is to estimate the proportions accurately; therefore, increasing the instance-level accuracy is not necessary.
We thus evaluated the instance-level accuracy of the training data.

\begin{figure}[t]
    \centering
    \begin{minipage}[b]{0.48\linewidth}
        \centering
        \includegraphics[width=\linewidth]{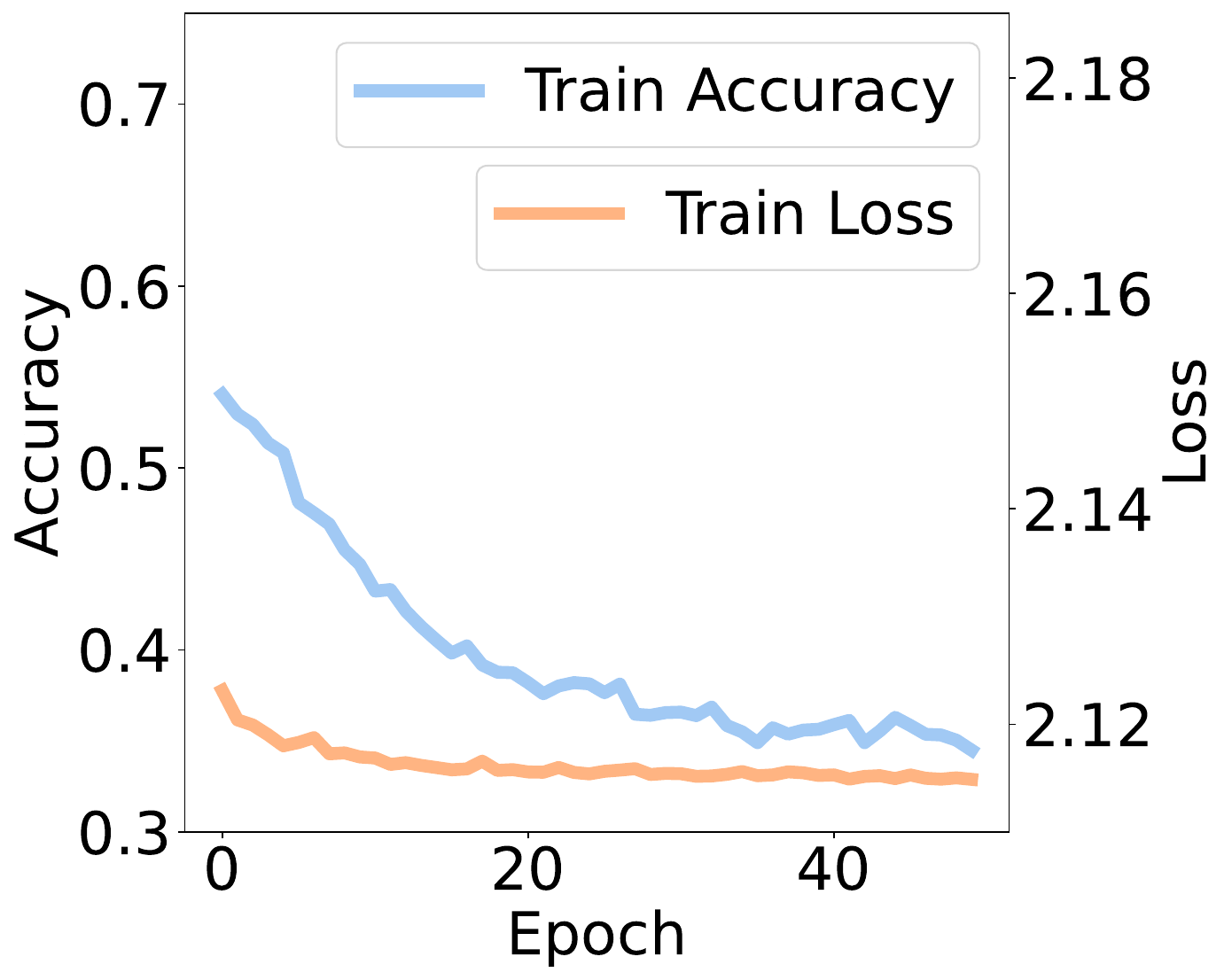}
    \end{minipage}
    \hfill
    \begin{minipage}[b]{0.48\linewidth}
        \centering
        \includegraphics[width=\linewidth]{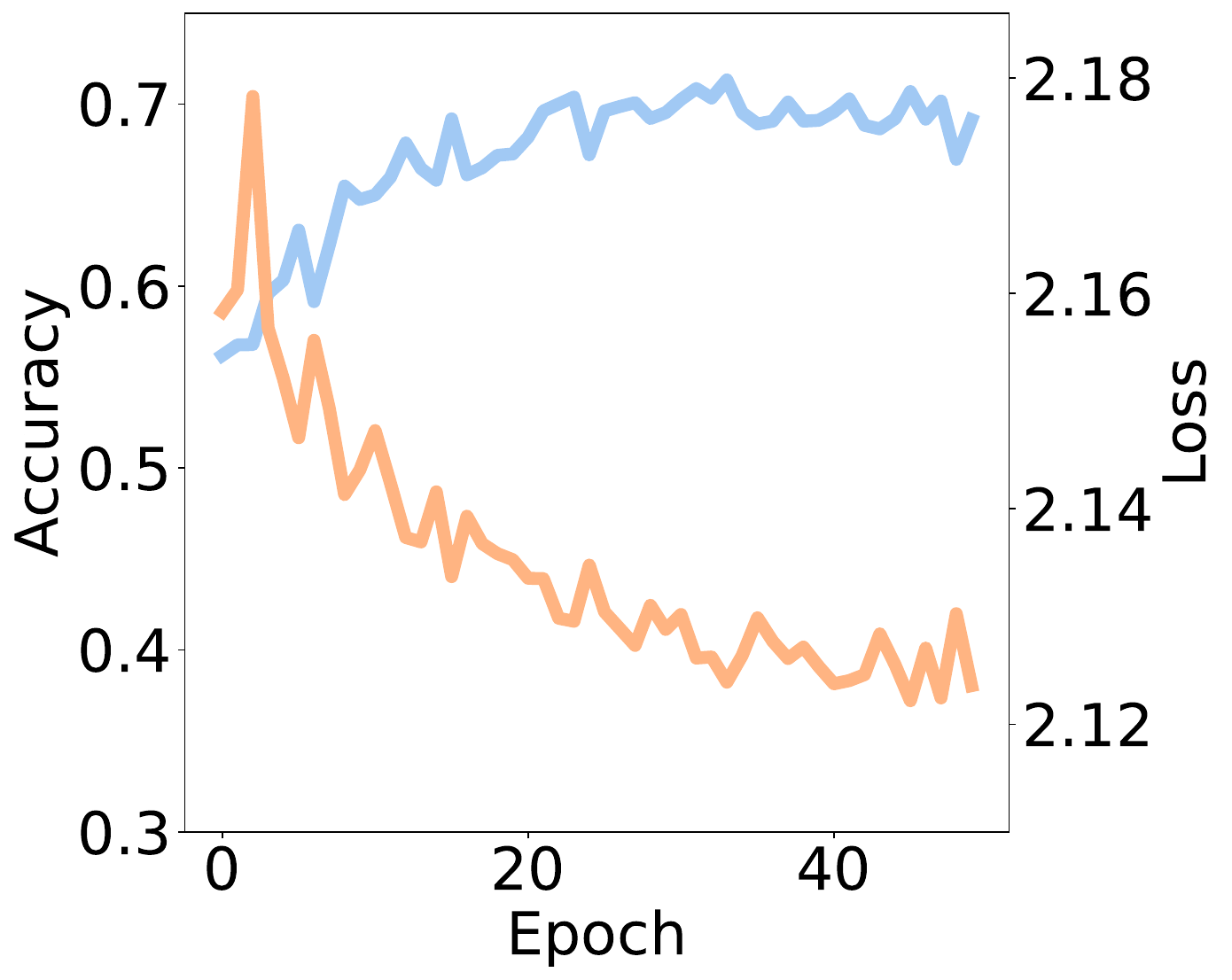}
    \end{minipage}
    \vspace{5pt}
    \caption{Training accuracy (blue) and loss (red) of the baseline method (left figure) and our proposed method (right figure).}
    \label{fig:train_acc}
    \vspace{20pt}
\end{figure}

Figure~\ref{fig:train_acc} shows the loss curve (orange) and instance-level accuracy (blue) for both the baseline method (Left) and the proposed method (Right) with a bag size of 12.
Although the proportion loss converged in the baseline method, the accuracy decreased with increasing epochs due to overfitting the noisy proportion labels.
In contrast, our method exhibits a decreasing loss with perturbations. We observe an improvement in accuracy without overfitting.

\begin{figure}[t]
    \centering
    \begin{minipage}[b]{0.48\linewidth}
        \centering
        \includegraphics[width=\linewidth]{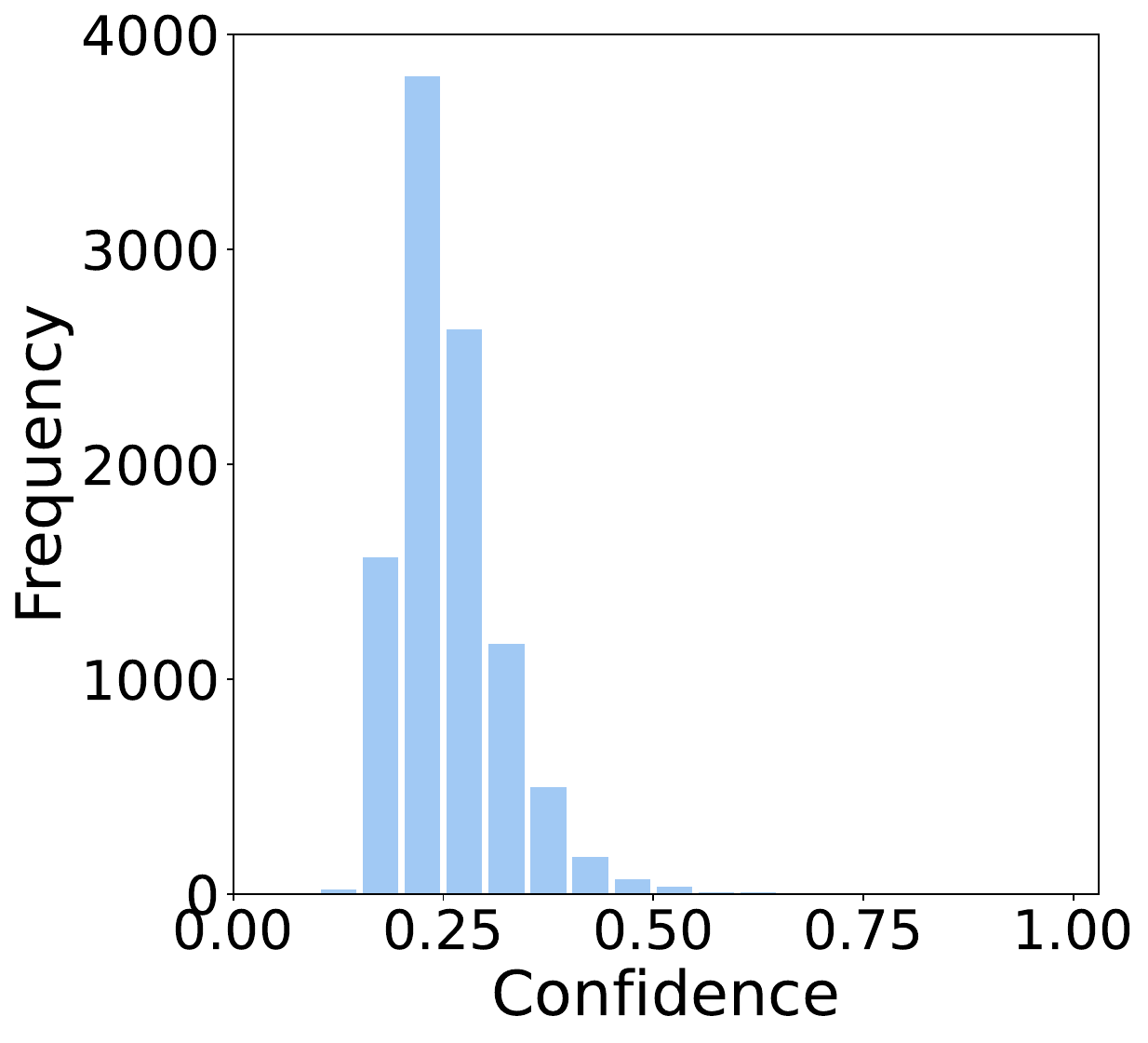}
    \end{minipage}
    \hfill
    \begin{minipage}[b]{0.48\linewidth}
        \centering
        \includegraphics[width=\linewidth]{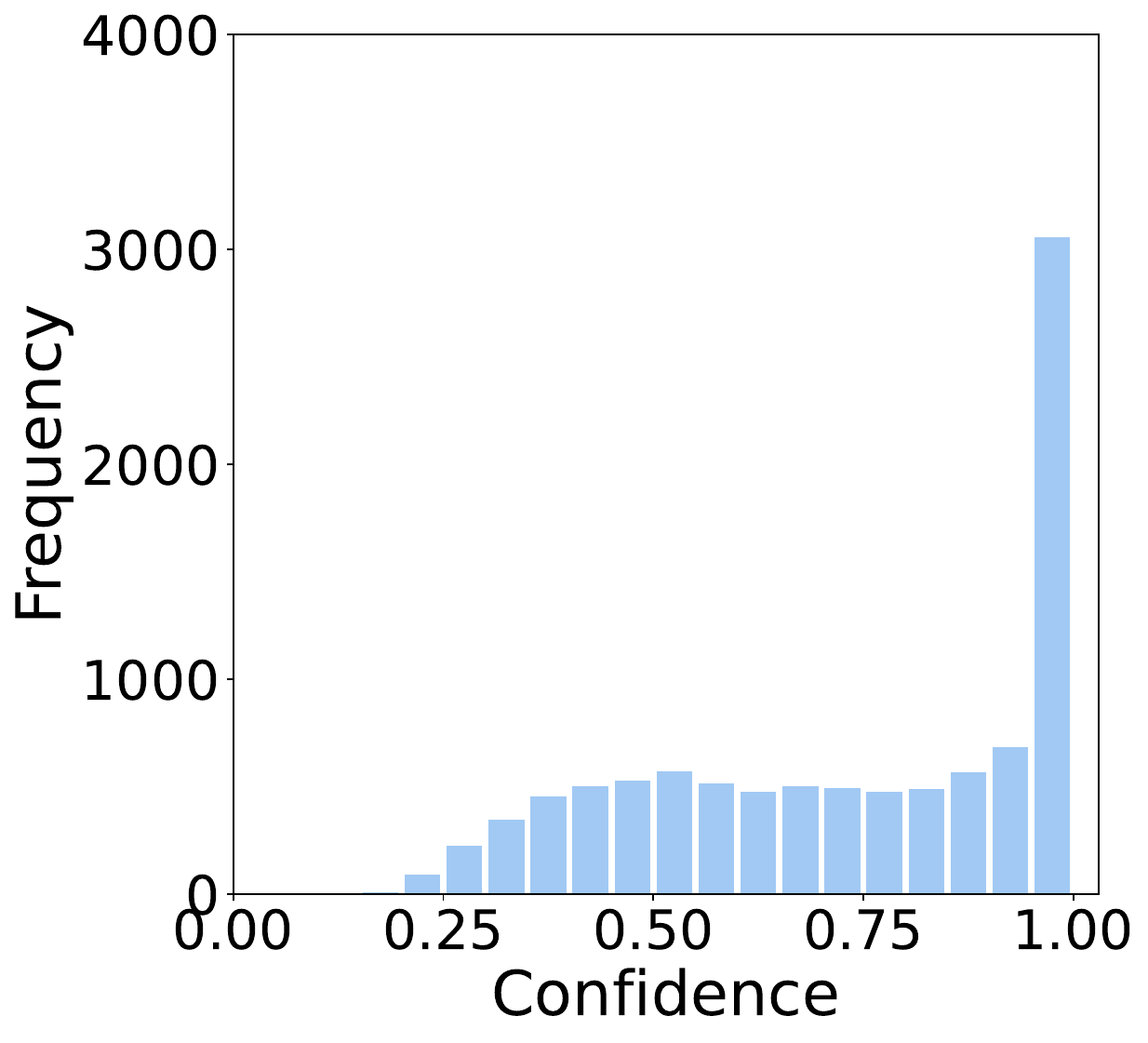}
    \end{minipage}
    \vspace{5pt}
    \caption{Histogram of confidence scores from the neural network output using the baseline method (left figure) and the proposed method (right figure).}
    \label{fig:confidence_histogram}
    \vspace{20pt}
\end{figure}

\begin{figure}[t]
\vspace{3mm}
    \centering
    \begin{minipage}[b]{0.48\linewidth}
        \centering
        \includegraphics[width=\linewidth]{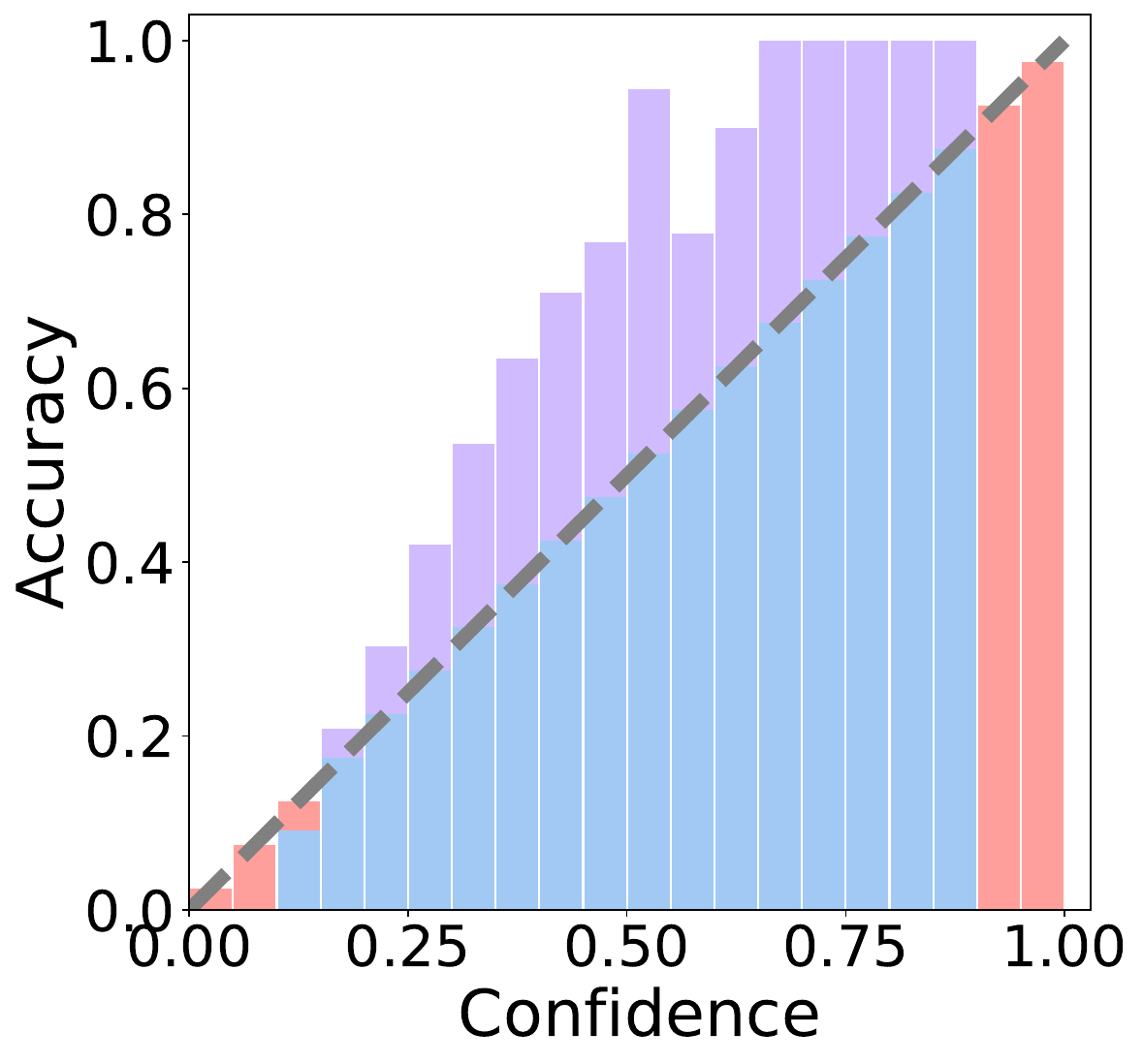}
    \end{minipage}
    \hfill
    \begin{minipage}[b]{0.48\linewidth}
        \centering
        \includegraphics[width=\linewidth]{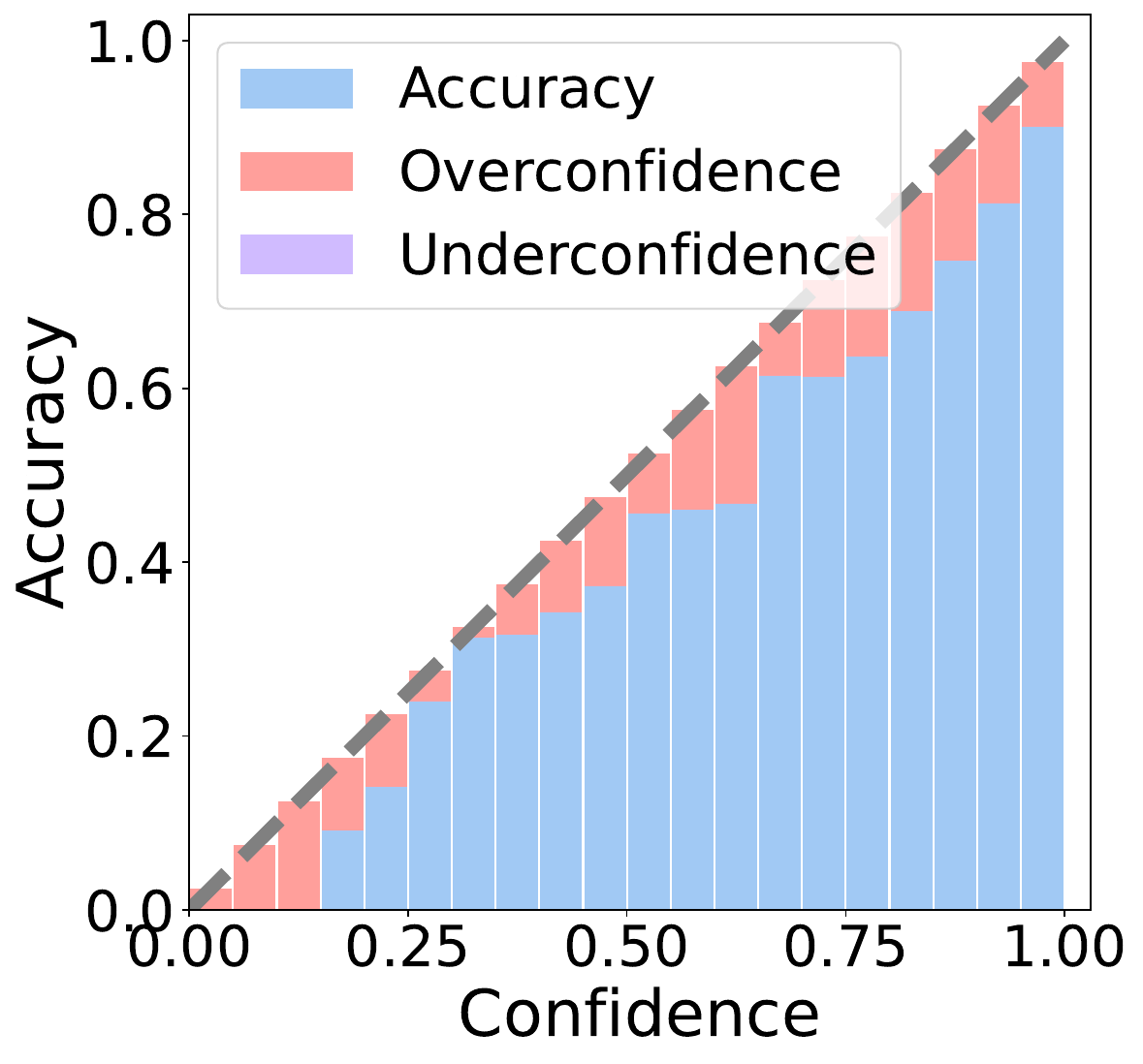}
    \end{minipage}
    \vspace{5pt}
    \caption{Relationship between prediction confidence and accuracy for the baseline method (left) and the proposed method (right). The dotted diagonal line represents perfect calibration, where confidence aligns perfectly with accuracy.}
    \label{fig:confidence_acc}
    \vspace{20pt}
\end{figure}

We examined the distributions of confidence to analyze the reasons behind the aforementioned results.
Figure~\ref{fig:confidence_histogram} illustrates the histograms of confidence for both the baseline (Left) and our method (Right).
In the baseline scenario (Left), confidence levels are generally low, predominantly clustered around 0.2 to 0.4. When confidence levels are low, even if the sum of confidences within a bag closely matches the expected proportion, the instance-level class labels may differ from the ground truth.
Consider a simple case of 3-class classification with a bag proportion of (0.00, 0.33, 0.67), a bag size of 3, and instance-level classes of 1, 2, and 2.
If the estimated confidences for the three instances are (0.00, 0.33, 0.67), (0.00, 0.33, 0.67), and (0.00, 0.33, 0.67), the sum of confidences closely matches the specified proportion. However, the instance-level classes are all labeled as 2.
In contrast, the proposed method yields high confidence, with 1 being the most frequently occurring value. If the confidence only takes 1, the solution space is limited, and thus, it tends to estimate correct class labels. Consider the above case, the estimated confidence only takes (0, 1, 0), (0, 0, 1), (0, 0, 1).
Therefore, the accuracy of the baseline method is worse than the proposed method.

Figure~\ref{fig:confidence_acc} shows the relationship between estimated confidence and accuracy. The diagonal lines represent points where confidence and accuracy are equal. The accuracy shown is the average accuracy within a confidence interval of 0.05. The purple (underconfidence) and red (overconfidence) regions highlight the gaps from the ideal accuracy (the diagonal line).
The baseline deviates from the diagonal line, with a tendency for actual accuracy to exceed confidence. In contrast, our proposed method closely aligns with the diagonal line, indicating better confidence calibration than the baseline method.

\section{Effectiveness on clinical data}
To demonstrate the effectiveness of our method on clinical data, we applied it to pathological images.

In some pathological diagnoses, the disease grade is determined based on the proportion of cancer subtypes in whole slide images (WSIs), with this proportion often recorded as diagnostic information in clinical settings.
Due to the enormous size (e.g., 100,000 × 50,000 pixels) of WSIs, they cannot be directly inputted into a CNN. Consequently, the majority of methods adopt a patch-based classification strategy, which involves segmenting the large image into smaller patches.
Therefore, this problem can be considered an LLP problem: a WSI is a bag where each bag consists of a set of patch images (instances), and the label proportion is attached to a WSI.

\vspace{0.5\baselineskip}
\noindent
{\bf Dataset of Chemotherapy:}
The dataset of chemotherapy contains three classes: non-tumor (negative), tumor-bed (positive), and residual-tumor (positive). In clinical practice, the viable tumor rate is diagnosed by calculating the ratio of the residual tumor region size to the sum of the tumor bed and residual tumor regions. Thus, the label proportion has been recorded as the diagnosis information.
The dataset consists of 143 cases of WSIs collected at Kyoto University Hospital. The size of each WSI is approximately $70000 \times 50000$ pixels, and each WSI was divided into $256 \times 256$ pixel patches to form a bag, where the bag sizes range from 700 to 2000.
Due to the GPU memory limitation for training, the maximum sampling number is 64 on RTX A6000 (48GB). So, we evaluated two different sample sizes, 32 and 64.

\vspace{0.5\baselineskip}
\noindent
{\bf Experimental setup:}
The network and its training setup remain consistent with those described in Section 6.1. We conducted 5-fold cross-validation for evaluation, wherein patients were partitioned into training, validation, and testing sets, ensuring that images from the same patient did not appear in different sets.
The batch size was set to 8 bags for training.

We used mean Dice (mDice) of instance-level classes for evaluation instead of accuracy since the dataset is class imbalanced.

\begin{table}[t]
\centering
\fontsize{10pt}{12pt}\selectfont
\begin{tabular}{l|cc}
\hline
Sample size & 32 & 64 \\
\hline
PL & 0.7847 & 0.8129 \\
LLPFC-uniform & 0.4387 & 0.4387 \\
LLPFC-approx & 0.3957 & 0.3957 \\
OPL & 0.3689 & 0.3689 \\ \hline
Ours & \bf{0.8059} & \bf{0.8168} \\
\hline
\end{tabular}
\vspace{2mm}
\vspace{5pt}
\caption{mDice of each method on Chemotherapy. Note that LLPFC and OPL are pseudo-labeling-based methods, and thus, they do not require sampling to make mini-bags. Therefore, the mDice is completely the same with both sample sizes.}
\label{tab:rinsho}
\vspace{10pt}
\end{table}

\vspace{0.5\baselineskip}
\noindent
{\bf Comparative study:}
To show the effectiveness of our method, we evaluated our method with four comparative methods: 1) proportion loss (PL) as a baseline, which uses the proportion of the original big bag as supervision for the sampled mini-bag.
2) LLPFC-uniform and 3) LLPFC-approx, in which LLPFC~\cite{Zhang2022llpfc} is the state-of-the-art of LLP methods. This method gives pseudo labels for each instance based on the noise transition matrix and has two types of approaches: ``uniform'' and ``approximate''.
4) Online Pseudo-Label Decision (OPL)~\cite{matsuo2023opl}, which is the state-of-the-art method designed for large bag in LLP. This method also gives pseudo labels for each instance based on the regret minimization
framework, which is one of the theories for online decision-making.
These three methods can avoid memory limitation by adding pseudo labels.

Table~\ref{tab:rinsho} shows the mDice of comparative methods on Chemotherapy. Although LLPFC and OPL can learn without the need for sampling, the performance of the initial pseudo-labels was poor on this dataset, and the learning did not progress well. Therefore, their performance is worse than the baseline method.
Our method outperformed the comparative methods.

\vspace{0.5\baselineskip}
\noindent
{\bf Effectiveness of the multivariate hypergeometric distribution:}
To demonstrate the effectiveness of the theoretically designed perturbation, we conducted an ablation study by substituting the multivariate hypergeometric distribution with a Gaussian distribution. Unlike our method, which automatically determines the hyperparameters of a distribution based on the proportion of the original bag, the size of the original bag, and mini-bags, the Gaussian distribution requires the specification of a hyperparameter: the standard deviation (sd). Consequently, we evaluated the instance-level classification performance using different sd values: 0.05, 0.15, and 0.25.

We evaluated these methods using two datasets: CIFAR-10 (with a sample size of 50) and Chemotherapy (with a sample size of 32), with accuracy being used as the performance metric for CIFAR-10, and mDice for Chemotherapy.

The results are presented in Table~\ref{tab:gaussian}.
The improvements achieved by the method incorporating perturbation with Gaussian distribution were limited across all standard deviation (sd) values.
This limitation arises because the perturbation was based on a fixed Gaussian distribution with constant parameters (standard deviation) for all bags, which may not be suitable for individual bags.
In contrast, our method automatically determines the distribution parameters based on the bag’s proportion, size, and sampling size. Due to this advantage, our method consistently outperformed methods using fixed Gaussian distribution.

\begin{table}[t]
\centering
\fontsize{10pt}{12pt}\selectfont
\begin{tabular}{l|c|c}
\hline
Dataset & \multicolumn{1}{c|}{CIFAR-10} & \multicolumn{1}{c}{Chemotherapy} \\ \hline
Evaluation metric & \multicolumn {1}{c|}{Accuracy} & \multicolumn{1}{c}{mDice} \\ \hline
 Sample size & 50 & 32\\
\hline
PL  & 0.4915 & 0.7847 \\
PL (Gaussian sd 0.05) & 0.5131 & 0.7813 \\
PL (Gaussian sd 0.15) & 0.5956 & 0.7557 \\
PL (Gaussian sd 0.25) & 0.5912 & 0.7166 \\ \hline
Ours & \bf{0.6401} & \bf{0.8059} \\
\hline
\end{tabular}
\vspace{2mm}
\vspace{5pt}
\caption{Accuracy and mDice of each method on CIFAR-10 and Chemotherapy.}
\label{tab:gaussian}
\vspace{20pt}
\end{table}

\section{Conclusion}
In this study, we propose a method to address the large bag size issues in LLP (Large Bag Problems), where the large number of images in a bag cannot be inputted into a network for training due to GPU memory limitations. To tackle this issue, we use a sampling method to generate mini-bags from the original bags. However, the proportion of images in a mini-bag is unknown. We model this proportion using a multivariate hypergeometric distribution. Additionally, we propose a perturbation method that applies different proportions in each iteration of training, in conjunction with the multivariate hypergeometric distribution. We also introduce a loss weighting technique that leverages this distribution. The proposed method helps mitigate overfitting to noisy proportions. Experimental results demonstrate that our method is effective and outperforms state-of-the-art techniques. Addressing large bag sizes in LLP under computational constraints has significant implications for researchers in the field.

We believe that this perturbation approach can be extended to address noisy labels in LLP problems. Proportions are often subjectively diagnosed in clinical, and therefore, the given proportions may contain noise. In future work, we plan to extend our perturbation approach to handle noisy proportion labels.





\begin{ack}
This work was supported by SIP-JPJ012425, JSPS KAKENHI Grant JP23K18509, JSPS-JP23KJ1723, and JST-JPMJAX23CR.
\end{ack}

\bibliography{main}

\end{document}